# Synthetic Industrial Object Detection: GenAI vs. Feature-Based Methods

Jose Moises Araya-Martinez[a,b,*], Adrián Sanchis Reig[a], Gautham Mohan[a], Sarvenaz Sardari[a], Jens Lambrecht[c] and Jörg Krüger[b]

[a]Mercedes-Benz AG, Future Automotive Manufacturing, Benz-Str. Bau40, 7963 Sindelfingen, Germany
[b]Technical University Berlin, Industrial Automation Technology, Pascalstraße 8-9, 9587 Berlin, Germany
[c]Technical University Berlin, Industry Grade Networks and Clouds, Ernst-Reuter-Platz 7, 9587 Berlin, Germany

* Corresponding author. E-mail address: jose_moises.araya_martinez@mercedes-benz.com

**Abstract**

Reducing the burden of data generation and annotation remains a major challenge for the cost-effective deployment of machine learning in industrial and robotics settings. While synthetic rendering is a promising solution, bridging the sim-to-real gap often requires expert intervention. In this work, we benchmark a range of domain randomization (DR) and domain adaptation (DA) techniques, including feature-based methods, generative AI (GenAI), and classical rendering approaches, for creating contextualized synthetic data without manual annotation. Our evaluation focuses on the effectiveness and efficiency of low-level and high-level feature alignment, as well as a controlled diffusion-based DA method guided by prompts generated from real-world contexts. We validate our methods on two datasets: a proprietary industrial dataset (automotive and logistics) and a public robotics dataset. Results show that if render-based data with enough variability is available as seed, simpler feature-based methods, such as brightness-based and perceptual hashing filtering, outperform more complex GenAI-based approaches in both accuracy and resource efficiency. Perceptual hashing consistently achieves the highest performance, with mAP50 scores of 98% and 67% on the industrial and robotics datasets, respectively. Additionally, GenAI methods present significant time overhead for data generation at no apparent improvement of sim-to-real mAP values compared to simpler methods. Our findings offer actionable insights for efficiently bridging the sim-to-real gap, enabling high real-world performance from models trained exclusively on synthetic data.

## 1. INTRODUCTION

In recent years, the development of machine learning models has seen significant advancements across various industries. However, these advancements often rely on large, annotated datasets, whose acquisition and annotation often implies prohibiting time overheads [1]. This challenge is particularly evident in industrial applications, where the complexity and variability of real-world environments, as well as the vast number of applications make it difficult to obtain sufficient labeled data for training robust models [2, 3, 4].

Many approaches are being developed to tackle this problem. For instance, data augmentation has emerged as a crucial technique to mitigate this challenge by artificially expanding training datasets through transformations of existing data [5]. While traditional augmentation methods, such as image rotation, flipping, cropping [5] and noise addition, are effective in increasing data diversity, they often fail to capture the full complexity of real-world scenarios [6]. Moreover, these methods do not fully address closing the sim-to-real gap, where models trained on simulated or augmented data struggle to generalize effectively to real-world environments [4, 6].

GenAI provides a promising solution to these limitations by employing deep learning models to create synthetic data that closely resembles real-world conditions, while also enhancing randomization [7]. Advancements in generative models, such as Stable Diffusion [7], Conditional Diffusion models [7, 8], and 3D simulation-based rendering frameworks [9, 10], have shown potential in generating high-fidelity synthetic data [6, 11, 12, 13]. These models not only enhance data diversity but also enable control over specific attributes, making it possible to create more realistic and varied datasets that better mimic real-world scenarios [7, 8].

This paper investigates the effectiveness of generative AI techniques in **industrial data augmentation for DR and DA** [14, 15], comparing them to other approaches in **narrowing the sim-to-real gap**. Specifically, we conduct a comparative study of six leading approaches: 1) Blender-based synthetic dataset of domain randomized scenes, 2) DA based on perceptual hashing, 3) DA based on low-level features (brightness), 4) Stable Diffusion XL with depth and canny conditioning combined with an IP-adapter for DA by following a background augmentation scheme with domain-aware prompts, and lastly, the same conditioned Stable Diffusion XL model with random prompts for enhanced data diversity starting from a rendered dataset.

By benchmarking these methods on proprietary as well as public industrial public datasets, we assess their ability to efficiently generate real-world-relevant data, for high model performance in real-world applications.

The contributions of this paper are threefold:
- **Comparative Analysis:** We provide a comprehensive comparison of multiple DR and DA methods, including generative AI techniques, in the context of synthetic industrial data generation and augmentation.
- **Sim-to-Real Gap Evaluation:** We investigate the impact of each method towards reducing the sim-to-real gap by evaluating their effectiveness in enhancing model generalization to real-world environments. Also, we evaluate the efficiency of each method to adapt to new applications by measuring the total generation and training time to obtain a certain mAP value.
- **Practical Insights:** Our results offer practical insights and recommendations for deploying these techniques in industrial machine learning pipelines, with a focus on reducing data burdens and improving real-world performance, especially for time-sensitive applications.

## 2. Related Works

This section offers an overview of the state of the art that inspired our work. We overview GenAI- and render-based data generation methods and sim-to-real techniques.

### 2.1. Synthetic Data Generation via Generative Models and Rendering engines

Generative models like GANs and diffusion models have gained prominence as alternatives to traditional data augmentation methods. GANs, introduced by Goodfellow et al. [16], generate realistic synthetic data by learning the distribution of dataset and creating new examples. Diffusion models, such as those by Ho et al. [17], have emerged as powerful alternative to GANs, generating images by gradually transforming noise into coherent data. With conditional controls, e.g. text prompts, depth maps and canny edges [8], they provide fine-grained control over the generated outputs. While diffusion models have shown success in creative domains, their application to industrial data augmentation and sim-to-real transfer is still developing. Recent studies [18, 19] show promise in improving datasets for tasks like classification, but further research is required to validate their effectiveness in industrial applications.

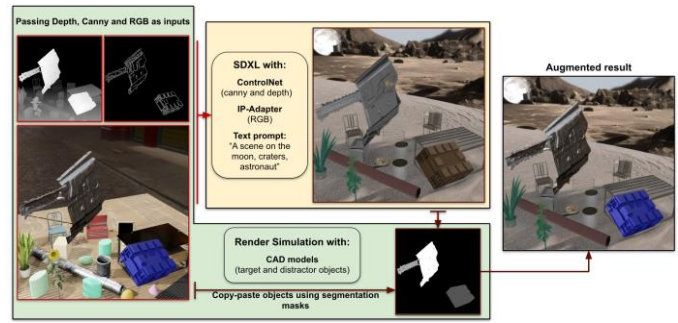

Fig. 1. Diagram of our GenAI pipeline using SDXL with ControlNet and IP-adapter for generating a new background with DA.

Furthermore, the Stable Diffusion Inpainting Pipeline [20] has been evaluated to automate the creation and annotation of object-detection datasets [21], however, some issues such as loose bounding boxes and data diversity have still to be addressed to become a viable option to automate creation of industrial datasets [21].

3D simulation platforms can be used to generate synthetic data that closely mimics real-world conditions. Blender [9], as a versatile 3D creation suite, enables the creation of large-scale synthetic datasets through high-fidelity 3D simulations. Previous studies have demonstrated the potential of simulated environments in reducing the sim-to-real gap in single-target object detection by training on 2500 – 5700 synthetic images [11, 4]. Newer approaches have reduced the required number of synthesized images by a factor of ten by achieving higher adaptation via low- and high-level feature matching between the simulated and real domains [12]. Furthermore, techniques aim at generating higher value data via eXplainable AI and latent space affinity assessments [13]. However, the trade-offs between synthetic data relevance and real-world performance on object detection are still active areas of research.

### 2.2. Sim-to-Real Transfer

Sim-to-real transfer focuses on enabling models to generalize from simulated to real-world environments. Approaches like DR [14] are used to bridge this gap. Domain randomization trains models on diverse simulated environments for robustness, while DA aligns simulated and real data distributions [22].

Although significant progress has been made in data augmentation, generative models, and sim-to-real transfer, the intersection of these areas in industrial contexts is still an emerging field. Achieving seamless transfer in complex industrial settings remains challenging, especially as the generation and training time to bring up new applications can impact production downtime.

### 3. Methodology

Following previous approaches by Araya-Martinez et al. [12, 13], we employ available computer-aided design (CAD) files to create a dataset ($D_{train}$) representing objects of interest with high fidelity with respect to shapes, perspectives and textures, as shown in Fig. 2(a). Additionally, as shown in Figs. 2(b) and 2(c), we split real-world images into a train set ($D_{train}$) and a reference set ($D_{ref}$). Our methodology and experiments consider the following points:

- **Rendering engine as baseline:** as in previous approaches, we assume the availability of CADs of the objects of interest in production and industrial domains [4, 12, 13, 11]. Thus, we aim at exploiting their usage for synthetic data generation as a first step towards sim-to-real gap narrowing.
- **GenAI vs. feature-based for sim-to-real gap reduction:** It has been demonstrated that sim-to-real gap narrowing by domain randomization using rendering engines is possible for object detection tasks using a large amount of training images [11]. We aim at incorporating the usage of deep-generative models to achieve sim-to-real gap narrowing on a multi-target object detection task and compare it against feature-based methods [12].
- **Minimal data acquisition and annotation:** Acquisition and annotation of real images is costly and time consuming. Thus, we aim at creating a dataset-generation pipeline with methods that minimize manual acquisition and annotation.
- **Resource and energy efficiency:** Deep Generative models such as Stable Diffusion are time and energy hungry at train and inference times [23]. For this reason, we compare their generation time against more efficient methods.

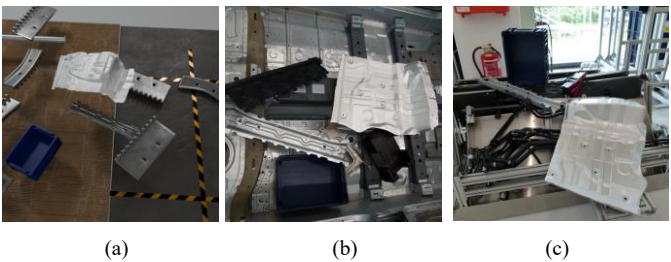

    (a)                    (b)                    (c)

Fig. 2. (a) rendered image from our proprietary dataset ∈ $D_{train}$; (b) sample test image ∈ $D_{test}$; (c) reference real image ∈ $D_{ref}$.

#### 3.1. Render-based Domain Randomization: Baseline Dataset

Our study employs two baseline datasets composed of synthetic images rendered with Blender's Cycles rendering engine [10]. The first dataset contains proprietary data conformed by three different target objects relevant to the automotive and logistics industries [12]. The second dataset is publicly available [24] and comprises 10 different industrial pieces from a robotics setting. The resulting baseline synthetic datasets contain 2000 rendered images each and were created with domain randomization techniques in a controlled scene. The ranges of variation of the multiple variables involved object-to-camera poses, lighting blur, and other variables of interest were set by a simulation expert in accordance with realistic parameters found in the real world [12]. We randomly split these images into a train ($D_{train}$) (80%) and validation ($D_{val}$) (20%) sets. This initial dataset also provides automated annotations of masks, poses, surface normals and bounding boxes, which are useful for further data-diversification steps. Both (proprietary and public) datasets contain real images which are split into a $D_{test}$ used to test the real-world performance of the approaches and a $D_{ref}$, which we use for sim-to-real augmentation and adaptation.

#### 3.2. Domain Adaptation

Automated DA techniques use information from the target context in form of pixel contents (low-level features) or semantic information (high-level features) to help narrow down the sim-to-real gap at a cost of generalization [11, 25].

To select the most relevant synthetic images based on real-world features, we acquire real images from $D_{ref}$. Based on this context information, we benchmark DA techniques as follows:

- **Low-level Features:**
  - Brightness-based image selection (Eq. 1).
- **High-level Features (semantic information):**
  - Perceptual Hashing [26] and Hamming distance [30].
  - Stable Diffusion XL with ControlNet (depth, canny) and IP-adapter [20].

*1) Brightness-Based Image Selection:* We explore DA based on low-level features by filtering images in $D_{train}$ by its brightness-similarity to images in $D_{ref}$. Brightness is calculated as the average normalized pixel value across each image [27]. Let **I** be a grayscale image with *M* rows and *N* columns. The pixel value at pixel (*i*, *j*) is denoted by $I_{i,j}$, and $I_{max}$ represents the maximum pixel value. Thus, the brightness B of the image is given by:

$$B_{brightness} = \frac{1}{M \cdot N} \sum_{i=1}^{M} \sum_{j=1}^{N} \frac{I_{i,j}}{I_{max}} \quad (1)$$

*2) Perceptual Hashing:* Perceptual hashing allows to represent the semantic information of an image as a single hash number with a desired bit-depth [28]. The operations involved in the extraction of the hash numbers are designed to preserve the image's hash representation even in presence of changes in the pixel-space such as brightness, cropping and scaling [29].

Then, the semantic content of two images can be compared by calculating the Hamming Distance [30] between their binary hash arrays, as shown in Eq. 2. For instance, let *A* and *B* be the perceptual hash numbers of two arbitrary images. Let *n* represent the bit-length of *A* and *B*. We computed the Hamming distance $\Delta_{Hamming}$ using the cardinality operator, by counting the number of binary elements that are different in each hash array, according to:

$$\Delta_{Hamming} = |\{ i \in \{1, \ldots, n\} \mid A_i \neq B_i \}| \quad (2)$$

3) *Stable Diffusion XL with conditioning:* We also evaluated using Stable Diffusion XL with ControlNet as DA technique for keeping the scene composition following depth and canny edge conditions [20] along with an IP-adapter. For convenience, we use the Hugging Face implementation [20]. During the generation process, depth data is obtained as part of the annotations generated with Blender, while canny edges information is obtained from OpenCV's canny edge detector [31]. Both conditions are forwarded to the hugging face's Stable Diffusion XL (SDXL) pipeline as additional channels to be used along with a prompt for image generation. Moreover, the original RGB pixel contents are added as well to the pipeline through the IP-Adapter of the hugging face's implementation [20]. The use of both ControlNet conditions along with the IP-Adapter ensures that the original composition of the input image is maintained, and the objects of interest can be copied from the original image onto the generated image to avoid undesired object modifications, as shown in Algorithm 1 and illustrated in Fig. 1 and 3. A changing random seed and the added conditions ensures that no two images are treated the same way by the pipeline, ensuring background randomization. It is worth noting that this approach does not require any real image, only high-level contextualized prompts to carry out the image generation under the DA concept. As shown in Fig. 1, we implement automated DA by obtaining the contextualized prompts from an image-to-text pipeline which receives the images $D_{ref}$ and returns a high-level caption of them [32]. An example of an automatically contextualized prompt is:

**Prompt:** "there is a desk with a monitor, keyboard, and other electronics on it in a room with a lot of clutter"
**Negative Prompt:** "bad, deformed, ugly"

In Algorithm 1, the background in the final image *i'* (Fig. 3(c) and 3(f)) has changed following the prompt without altering the objects of interest. Here, we set the scale for all the conditions to 0.5, and the default values of the pipeline for the guidance scale and the denoiser (5 and 50 respectively).

*3.3. GenAI Domain Randomization*

This approach also relies on the Stable diffusion XL with ControlNet and IP-adapter pipeline [20]. However, it features the difference of not having context-aware prompts during each for-loop iteration of Algorithm 1. For instance, in our experiments, we use prompts such as:

- "A scene on the moon, craters, astronaut"
- "A dense forest with sunlight filtering through the trees"
- "A snow-covered mountain range with clear blue skies"
- "An underwater coral reef teeming with fish"
- "A peaceful meadow with wildflowers and tall grass swaying in the breeze"
- "A peaceful beach with waves gently lapping the shore"
- "A desert landscape with sand dunes and clear night sky"
- "A grassy hillside with grazing animals under a bright blue sky"

**Negative prompt:** "bad, deformed, ugly, abstract"

The prompts are designed to avoid the model generating any industrial background. Thus prevent any overlap between the dataset created through automated prompts and with the Gen-AI backgrounds from random prompts. This decision is meant to increase the diversity of data and reduce the dependency of context knowledge to generate data of a specific application. The final images along with the input are shown in Fig. 3.

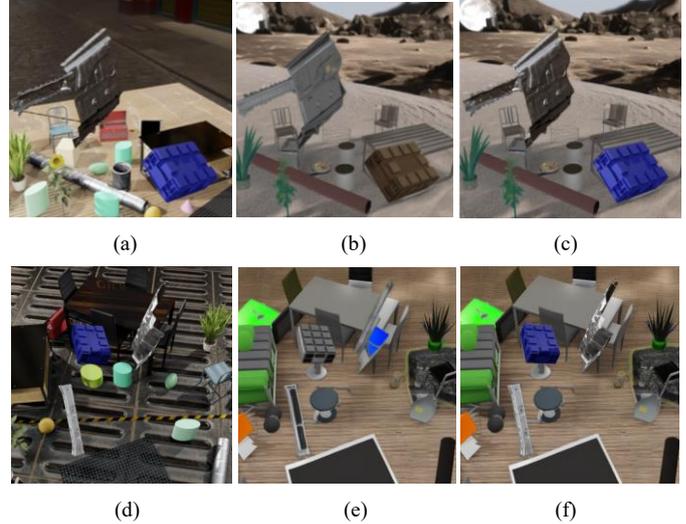

Fig. 3. Guided background randomization using SDXL with ControlNet, IP-Adapter. First row used random prompts: (a) render of proprietary dataset; (b) render after passing through the pipeline; (c) final output. Second row uses context-aware prompts with: (d) render of proprietary dataset; (e) render after passing through the pipeline; (f) final output.

---

**Algorithm 1** Background randomization using Stable Diffusion XL with ControlNet and IP-adapter

---

**Input:**
  The set of original images *I*.
  Depth maps of the original images *D*.
  Canny edges for the target objects in the original images *C*.
  Negative and positive text prompts.
**Output:** Images with same composition and target objects as the original images with a modified background.
**for** *i, d, c, m* **in** (*I, D, C, M*) **do**
  **Step 1:** Load the original image *i*.
  **Step 2:** Load the depth map *d*.
  **Step 3:** Load the canny edges *c*.
  **Step 4:** Load the binary masks *m*.
  **Step 5:** Initialize SDXL, ControlNet and IP-Adapter.
  **Step 6:** Define the textual negative and positive prompts.
  **Step 7:** Generate image *i'* using the pipeline with the image *i*, canny edges *c*, depth map *d*, prompt and negative prompt.
  **Step 8:** Paste the targets from the original image *i* onto *i'* using the binary mask *m*.
  **Step 9:** Save image *i'*.
**end for**

---

## 4. Experiments and Analysis

We evaluate the effectiveness of different domain randomization and domain adaptation methods by training the YOLOv8 model [33] on synthetic datasets generated by each method. YOLOv8 was chosen as its efficiency in both training and inference allows for faster experimentation and iteration. This makes it an ideal choice for benchmarking different data generation methods across multiple datasets [34, 33].

The model's performance was then tested on the real-world data $D_{test}$ for both the proprietary and public datasets, compared using the mAP at IOU=0.5 (mAP50) [35], as shown in Fig. 4.

The two plots in Figs. 4(a) and 4(b) show the mAP50 of YOLOv8 models trained with multiple number of images sampled up from the initial seed of 400 rendered images or augmented from the full dataset, containing 2000 images. Each line in the plot, except the "base_render" line which indicates the original rendered dataset, represents a specific data augmentation or filtering strategy as described in Section 3.1 and 3.2. For any method, the first 400 images are the same and every next 200 images added are based on an augmentation method. The data subsets created then undergo an 80:20 training-to-validation split.

In addition to the mAP values, we also monitor the time taken for generation of dataset and train the YOLO models which are given by the X axis of the plots. All timing values are generated in a workstation with a graphics card Nvidia RTX4090 and an Intel i9-14900K processor. The methods whose curves tend towards the upper left portion of the plots allow for faster augmentation and training and perform better at detection. Thus, exhibit a higher mAP-to-time efficiency.

It should be noted that filtering-based methods, such as perceptual hashing and brightness-based filtering, sort the original rendered dataset, comprising 2000 samples, and select the top N ranked images to add to the initial 400. In contrast, GenAI methods i.e. context–based prompting and random prompting apply generative augmentations to each subset of the original renders, creating new image variants for each training size.

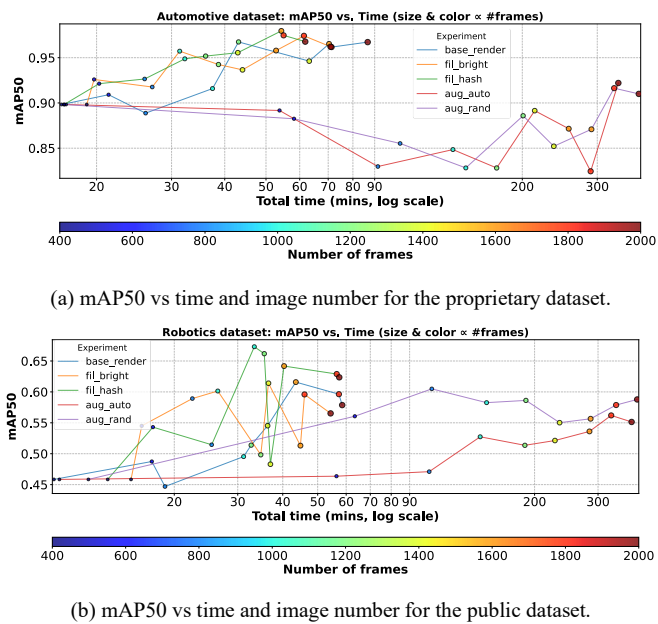

(a) mAP50 vs time and image number for the proprietary dataset.

(b) mAP50 vs time and image number for the public dataset.

Fig. 4. Benchmark of mAP50 values vs total time elapsed for data generation and model training (time overhead). The line colors represent data augmentation "aug" or filtering "fil" methods whereas the color bar represents the number of images involved in each training process. In (a) we show values for the proprietary automotive dataset and (b) values for the public robotics dataset.

For Fig. 4 (a), the result for the proprietary dataset indicates perceptual hashing as a clear winner, consistently outperforming every other method and peaking close to mAP 98%, while taking a total data generation and model training time of less than one hour. Brightness-based filtering follows this closely, while both filtering methods remain better performing than the original rendered images. The mAP50 values for GenAI based augmentation techniques are considerably lower, while also taking considerably more time for data augmentation. The peak performance by any GenAI method comes from the augmentation method with contextualized prompts at 92% for the proprietary dataset. However, it should be noted that in general random prompting yields better results, with a peak mAP50 of 91%.

The results of the public dataset illustrated in Fig. 4 (b) reflect a similar trend. This dataset has a higher multi-object detection complexity as it contains 10 objects with similar geometrical and texture characteristics. Under this new data distribution, perceptual hashing still yields better results, peaking at 67%, indicating its robustness across datasets. Moreover, it achieves data generation and model convergence of about 35min, which is three times less duration than the best performing GenAI method. The base rendered datasets gradually improve in mAP as more images are allowed into the train set but yield lower results when compared to any filtering method. As also observed in the proprietary dataset, the random-prompt-based GenAI method performs better than context aware prompting, peaking at nearly 58%.

When it comes to efficiency considerations, filtering-based methods perform better than GenAI based methods. They exhibit a lower total time overhead for total data preparation (rendering + filtering) and model training.

## 5. Conclusions and Future Works

In this work, we leverage Stable Diffusion's XL with depth and canny conditions and compare it against feature-based techniques in the context of sim-to-real gap narrowing. In our experiments we evaluate the effectiveness of a GenAI-based data augmentation pipeline against low-level feature-filtering techniques based on brightness and higher-level feature matcher based on perceptual hashing. Such techniques are benchmarked in the context of DR and DA by comparing their mAP and time overhead in multi-target proprietary and public industrial object detection datasets using YOLOv8.

In general, perceptual hashing emerged as a robust method for filtering relevant images. It exhibits the best trade-off between mAP values and time overhead required to filter data and train models. Our experiments indicate state-of-the-art results with mAP50 values of 98% and 67% for the proprietary and public datasets, respectively, while training on synthetic data selected with perceptual hashing DA.

Conversely, it was observed that GenAI-based augmentation did not significantly contribute to the final mAP. The investigated random prompt based GenAI method proved to outperform the context-aware prompting. Although counter intuitive, it might be explained by further investigation regarding the quality of context aware prompts.

Our findings suggest that an initial large, randomized dataset from a rendering engine and subsequent context-filtering with perceptual hashing or brightness offers a viable solution to efficiently close the sim-to-real gap in industrial settings. We envision further research in prompt engineering, testing on further datasets and intensifying the search for optimal generation parameters as enablers to unleash further potential of GenAI-based synthetic data generation.

**References**

[1] Nowruzi FE, Kapoor P, Kolhatkar D, Hassanat FA, Laganière R, Rebut J. How much real data do we actually need: analyzing object detection performance using synthetic and real data. 2023 [cited 2025 Jun 6]. Available from: https://arxiv.org/abs/2305.16289

[2] Demlehner Q, Laumer S. Shall we use it or not? Explaining the adoption of artificial intelligence for car manufacturing purposes. 2020 [cited 2025 Jun 6].

[3] Hansen EB, Bøgh S. Artificial intelligence and internet of things in small and medium-sized enterprises: a survey. J Manuf Syst 2021;58:362-372.

[4] Mayershofer C, Ge T, Fottner J. Towards fully-synthetic training for industrial applications. In: Proceedings of LISS 2020. Singapore: Springer; 2021. p. 765-782.

[5] Takahashi R, Matsubara T, Uehara K. Data augmentation using random image cropping and patching for deep CNNs. IEEE Trans Circuits Syst Video Technol 2020;30(9):2917-2929.

[6] Chebotar Y, Handa A, Makoviychuk V, Macklin M, Issac J, Fox D. Closing the sim-to-real loop: adapting simulation randomization with real-world experience. In: Proceedings of Robotics: Science and Systems (RSS). Virtual; 2021.

[7] Rombach R, Blattmann A, Lorenz D, Esser P, Ommer B. High-resolution image synthesis with latent diffusion models. In: Proceedings of the IEEE/CVF Conference on Computer Vision and Pattern Recognition (CVPR); 2022 Jun; New Orleans, LA, USA. p. 10684-10695.

[8] Zhang L, Rao A, Agrawala M. Adding conditional control to text-to-image diffusion models. In: Proceedings of the IEEE/CVF Conference on Computer Vision and Pattern Recognition (CVPR); 2023 Jun; Seattle, WA, USA. p. 1234-1243.

[9] Blender Foundation. Blender 4.0. 2023 [cited 2025 Jun 6]. Available from: https://www.blender.org/download/releases/4-0/

[10] Blender Foundation. The Cycles render engine. 2023 [cited 2025 Jun 6]. Available from: https://projects.blender.org/blender/cycles.git

[11] Eversverg L, Lambrecht J. Generating images with physics-based rendering for an industrial object detection task: realism versus domain randomization. Sensors 2021;21(23):7901.

[12] Araya-Martinez JM, Sardari S, Lambert M, Zak JA, Töper F, Krüger J, et al. A data-centric evaluation of leading multi-class object detection algorithms using synthetic industrial data. Procedia CIRP [in press] 2025.

[13] Araya-Martinez JM, Tom T, Sardari S, Sanchis Reig A, Mohan G, Shukla A, et al. Domain adaptation using vision transformers and XAI for fully synthetic industrial training. Procedia CIRP [in press] 2025.

[14] Tobin J, Fong R, Ray A, Schneider J, Zaremba W, Abbeel P. Domain randomization for transferring deep neural networks from simulation to the real world. In: Proceedings of the 2017 IEEE/RSJ International Conference on Intelligent Robots and Systems (IROS); 2017 Oct; Vancouver, BC, Canada. p. 23-30.

[15] Tremblay J, Prakash A, Acuña D, Brophy M, Jampani V, Anil C, et al. Training deep neural networks with synthetic data: bridging the reality gap by domain randomization. In: Proceedings of the IEEE/CVF Conference on Computer Vision and Pattern Recognition Workshops (CVPRW); 2018 Jun; Salt Lake City, UT, USA. p. 969-977.

[16] Goodfellow IJ, Pouget-Abadie J, Mirza M, Xu B, Warde-Farley D, Ozair S, et al. Generative adversarial nets. In: Proceedings of the Advances in Neural Information Processing Systems (NeurIPS); 2014 Dec; Montréal, Canada. p. 2672-2680.

[17] Ho J, Jain A, Abbeel P. Denoising diffusion probabilistic models. 2020 [cited 2025 Jun 6]. Available from: https://arxiv.org/abs/2006.11239

[18] Dunlap L, Umino A, Zhang H, Yang J, Gonzalez JE, Darrell T. Diversify your vision datasets with automatic diffusion-based augmentation. 2023 [cited 2025 Jun 6]. Available from: https://arxiv.org/abs/2305.16289

[19] Fang H, Han B, Zhang S, Zhou S, Hu C, Ye W-M. Data augmentation for object detection via controllable diffusion models. In: Proceedings of the 2024 IEEE/CVF Winter Conference on Applications of Computer Vision (WACV); 2024 Jan; Snowmass Village, CO, USA. p. 1246-1255.

[20] Hugging Face. ControlNet with Stable Diffusion XL. 2025 [cited 2025 Jun 6]. Available from: https://huggingface.co/CompVis/controlnet-sdxl

[21] Bellenfant T. Dataset Forge. 2023 [cited 2025 Jun 6]. Available from: https://github.com/bellenfanttyler/dataset-forge

[22] Ganin Y, Ustinova E, Ajakan H, Germain P, Larochelle H, Laviolette F, March M, Lempitsky V. Domain-adversarial training of neural networks. J Mach Learn Res 2016;17(59):1-35.

[23] Berthelot A, Caron E, Jay M, Lefevre L. Estimating the environmental impact of generative-AI services using an LCA-based approach. Procedia CIRP [in press] 2024.

[24] Horváth D, Erdős G, Szénási Z, Horváth T, Földi S. Object detection using sim2real domain randomization for robotic applications. IEEE Trans Robot 2023;39(4):1234-1242.

[25] Hodan T, Vineet V, Gal R, Shalev E, Hanzelka J, Connelly T, Kanouni B, Singh SN, Guenter B. Photorealistic image synthesis for object detection: bridging simulation and reality. In: Proceedings of the IEEE International Conference on Robotics and Automation (ICRA); 2023 May; London, UK. p. 842-849.

[26] Monga V, Banerjee A, Evans B. A clustering based approach to perceptual image hashing. IEEE Trans Inf Forensics Secur 2006;1(1):68-79.

[27] Encord. Encord Active. 2024 [cited 2025 Jun 6]. Available from: https://github.com/encord-team/encord-active

[28] Mihçak MK, Venkatesan R. New iterative geometric methods for robust perceptual image hashing. In: ACM Workshop on Digital Rights Management; 2001; Darmstadt, Germany. p. 13-21.

[29] Farid H. An overview of perceptual hashing. J Online Trust Safety 2021;1(1):1-22.

[30] Hamming RW. Coding and information theory. Englewood Cliffs, NJ: Prentice-Hall; 1986.

[31] OpenCV Developers. OpenCV: Canny edge detection. 2025 [cited 2025 Jun 6]. Available from: https://docs.opencv.org/4.x/da/d22/tutorial_py_canny.html

[32] Li J, Li D, Xiong C, Hoi SH. BLIP: Bootstrapping language-image pre-training for unified vision-language understanding and generation. 2022 [cited 2025 Jun 6]. Available from: https://arxiv.org/abs/2201.12086

[33] Jocher G, Chaurasia A, Qiu J. Ultralytics YOLO. 2023 Jan [cited 2025 Jun 6]. Available from: https://github.com/ultralytics/ultralytics

[34] Hussain M. Yolo-v1 to yolo-v8: the rise of YOLO and its complementary nature toward digital manufacturing and industrial deployment. Machines Tooling 2023 [cited 2025 Jun 6]. Available from: https://doi.org/10.1016/j.machinet.2023.100189

[35] Everingham M, Van Gool L, Williams CK, Winn J, Zisserman A. The PASCAL visual object classes (VOC) challenge. Int J Comput Vis 2009;88(2):303-338.